\documentclass{sigkddExp}
\usepackage[T1]{fontenc}
\usepackage[utf8]{inputenc}
\usepackage{booktabs}
\begin{document}
%
% --- Author Metadata here ---
% -- Can be completely blank or contain 'commented' information like this...
%\conferenceinfo{WOODSTOCK}{'97 El Paso, Texas USA} % If you happen to know the conference location etc.
%\CopyrightYear{2001} % Allows a non-default  copyright year  to be 'entered' - IF NEED BE.
%\crdata{0-12345-67-8/90/01}  % Allows non-default copyright data to be 'entered' - IF NEED BE.
% --- End of author Metadata ---

\title{A Survey of Scaling in Large Language Model Reasoning}
%\subtitle{[Extended Abstract]
% You need the command \numberofauthors to handle the "boxing"
% and alignment of the authors under the title, and to add
% a section for authors number 4 through n.
%
% Up to the first three authors are aligned under the title;
% use the \alignauthor commands below to handle those names
% and affiliations. Add names, affiliations, addresses for
% additional authors as the argument to \additionalauthors;
% these will be set for you without further effort on your
% part as the last section in the body of your article BEFORE
% References or any Appendices.

\numberofauthors{1}
%
% You can go ahead and credit authors number 4+ here;
% their names will appear in a section called
% "Additional Authors" just before the Appendices
% (if there are any) or Bibliography (if there
% aren't)

% Put no more than the first THREE authors in the \author command
%%You are free to format the authors in alternate ways if you have more 
%%than three authors.

\author{
\alignauthor Zihan Chen$^{1}$, Song Wang$^{1}$, Zhen Tan$^{2}$, Xingbo Fu$^{1}$, Zhenyu Lei$^{1}$, Peng Wang$^{1}$, Huan Liu$^{2}$, Cong Shen$^{1}$, Jundong Li$^{1}$ \\
\affaddr{$^{1}$University of Virginia, Charlottesville, VA, USA} \\
\affaddr{$^{2}$Arizona State University, Tempe, AZ, USA} \\
\email{\{brf3rx, sw3wv, xf3av, vjd5zr, pw7nc, cong, jundong\}@virginia.edu} \\
\email{\{ztan36, huanliu\}@asu.edu}
}
% \author{
% %
% % The command \alignauthor (no curly braces needed) should
% % precede each author name, affiliation/snail-mail address and
% % e-mail address. Additionally, tag each line of
% % affiliation/address with \affaddr, and tag the
% %% e-mail address with \email.
% \alignauthor Ben Trovato \\
%        \affaddr{Institute for Clarity in Documentation}\\
%        \affaddr{1932 Wallamaloo Lane}\\
%        \affaddr{Wallamaloo, New Zealand}\\
%        \email{trovato@corporation.com}
% \alignauthor G.K.M. Tobin\\
%        \affaddr{Institute for Clarity in Documentation}\\
%        \affaddr{P.O. Box 1212}\\
%        \affaddr{Dublin, Ohio 43017-6221}\\
%        \email{webmaster@marysville-ohio.com}
% \alignauthor Lars Th{\o}rv\"{a}ld\titlenote{This author is the
% one who did all the really hard work.}\\
%        \affaddr{The Th{\o}rv\"{a}ld Group}\\
%        \affaddr{1 Th{\o}rv\"{a}ld Circle}\\
%        \affaddr{Hekla, Iceland}\\
%        \email{larst@affiliation.org}
% }
% \additionalauthors{Additional authors: John Smith (The Th{\o}rvald Group,
% email: {\texttt{jsmith@affiliation.org}}) and Julius P.~Kumquat
% (The Kumquat Consortium, email: {\texttt{jpkumquat@consortium.net}}).}
% \date{30 July 1999}
\maketitle
\begin{abstract}
The rapid advancements in large Language models (LLMs) have significantly enhanced their reasoning capabilities, driven by various strategies such as multi-agent collaboration. However, unlike the well-established performance improvements achieved through scaling data and model size, the scaling of reasoning in LLMs is more complex and can even negatively impact reasoning performance, introducing new challenges in model alignment and robustness. In this survey, we provide a comprehensive examination of scaling in LLM reasoning, categorizing it into multiple dimensions and analyzing how and to what extent different scaling strategies contribute to improving reasoning capabilities. We begin by exploring scaling in input size, which enables LLMs to process and utilize a more extensive context for improved reasoning. Next, we analyze scaling in reasoning steps that improve multi-step inference and logical consistency. We then examine scaling in reasoning rounds, where iterative interactions refine reasoning outcomes. Furthermore, we discuss scaling in training-enabled reasoning, focusing on optimization through iterative model improvement. Finally, we outline future directions for further advancing LLM reasoning. By synthesizing these diverse perspectives, this survey aims to provide insights into how scaling strategies fundamentally enhance the reasoning capabilities of LLMs and further guide the development of next-generation AI systems.
\end{abstract}

\section{Introduction} % 1.5-page %czh
Large Language Models (LLMs) have evolved rapidly, demonstrating remarkable advancements across various natural language processing (NLP) tasks, including text generation, comprehension, and problem-solving~\cite{jiang2023active, sarthi2024raptor, xu2023recomp,xu2024chatqa, xu2023retrieval,jiang2024longrag}. One of the key driving forces behind these improvements is scaling, where increasing the size of training data and model parameters has led to substantial performance gains~\cite{lee2024inference,kaplan2020scaling,wei2022emergent}. Scaling has played a pivotal role in the development of state-of-the-art LLMs such as GPT-4~\cite{openai2024gpt4technicalreport}, and Gemini~\cite{team2024gemini}, enabling them to generalize across a broad range of tasks with unprecedented accuracy and fluency~\cite{Wang_2024}. The empirical success of scaling laws has reinforced the notion that simply increasing model size and data availability can significantly enhance LLM capabilities~\cite{chen2023meditron,muennighoff2023scaling,chung2024scaling}. However, while such scaling strategies have led to more powerful models, they do not fully explain improvements in complex reasoning tasks, which require structured thinking and logical consistency~\cite{du2023improving,saunshi2025reasoning,geiping2025scaling}.

Notably, unlike simpler tasks that rely on memorization or direct retrieval of information, reasoning demands deeper cognitive-like processes, including step-by-step deductions, counterfactual reasoning, and planning~\cite{paul2024refiner, kuo2025h}. While early LLMs exhibited shallow reasoning abilities~\cite{brown2020language,lu2021fantastically}, recent advancements have introduced techniques aimed at enhancing LLM reasoning performance through various strategies~\cite{deng2023implicit,hao2024training,shen2025efficient}.
For instance, s1~\cite{muennighoff2025s1} explicitly extends the reasoning length, enabling models to engage in deeper, iterative reasoning that can identify and correct errors in previous inference steps. However, scaling reasoning length does not always guarantee improved performance—simply increasing the number of reasoning steps may introduce redundancy, compounding errors, or even diminished accuracy~\cite{rasal2024llm, khan2024debating, michael2023debate, wu2025more, chen2024more, yang2025towards}. This highlights the complex and non-trivial nature of scaling in reasoning, necessitating a deeper investigation into how different scaling strategies influence LLM reasoning effectiveness and when they yield diminishing returns. In this survey, we use \textit{reasoning} to refer to tasks in which the model must perform nontrivial transformation over information, such as multi-step deduction or iterative refinement, rather than merely retrieve or fluently generate content. Under this view, not every improvement in general LLM capability should be interpreted as a reasoning improvement. For example, larger context windows, retrieval, or memory may improve performance simply by supplying missing evidence, while their reasoning benefit is most meaningful when that evidence must be integrated in a multi-step decision process. Conversely, these strategies may offer limited gains on tasks dominated by direct factual recall or shallow pattern matching. Throughout this survey, we focus on scaling strategies that strengthen reasoning behavior and highlight the task settings and failure modes under which their gains may diminish.

\begin{table*}[t]
\centering
\small
\setlength{\tabcolsep}{5pt}
\begin{tabular}{p{3.0cm}p{2.8cm}p{3.3cm}p{1.5cm}p{3.8cm}}
\toprule
\textbf{Survey} & \textbf{Scope} & \textbf{Organizing lens} & \textbf{Trade-offs} & \textbf{Main emphasis} \\
\midrule
Zhang et al.~\cite{zhang2025survey}
& Test-time scaling
& \emph{What, how, where, and how well} to scale
& Yes
& Taxonomy, assessment, and deployment guidance for \emph{test-time} scaling. \\

Ke et al.~\cite{ke2025survey}
& LLM reasoning
& \emph{Regimes} and \emph{architectures}, with input/output perspectives
& Partial
& Broad survey of inference scaling, learning to reason, and agentic systems. \\

Lai et al.~\cite{lai2025survey}
& Post-training scaling
& \emph{SFT, RLxF, and TTC} in post-training
& Partial
& Scaling after pre-training, especially alignment and post-training methodologies. \\

\textbf{Ours}
& Scaling in LLM reasoning
& \textbf{Input sizes, reasoning steps, reasoning rounds, and training-enabled reasoning}
& \textbf{Yes}
& \textbf{Reasoning-centric synthesis across scaling dimensions, with unified comparison of gains, costs, and failure modes.} \\
\bottomrule
\end{tabular}
\caption{Comparison with closely related recent surveys. Prior work focuses on test-time scaling, general reasoning regimes/architectures, or post-training scaling, while our survey emphasizes a unified, reasoning-centric view across multiple scaling dimensions.}
\label{tab:survey_positioning} 
\end{table*}

\begin{table*}[t]
\centering
\small
\begin{tabular}{p{2.8cm}p{2.3cm}p{2.2cm}p{2.2cm}p{2.5cm}p{2.0cm}}
\toprule
\textbf{Dimension} & \textbf{What is scaled} & \textbf{Main benefit} & \textbf{Main cost} & \textbf{Typical failure mode} & \textbf{Best-fit tasks} \\
\midrule
Input sizes & External information / context / demonstrations & Better grounding and task adaptation & Retrieval and long-context cost & Irrelevant context, distraction, lost-in-the-middle & Knowledge-intensive QA, long-context tasks \\
Reasoning steps & Intermediate reasoning depth & Better decomposition and verification & Higher token and search cost & Overthinking, compounding errors & Math, logic, planning \\
Reasoning rounds & Interaction rounds across agents/humans & Critique, diversity, iterative refinement & Coordination and latency overhead & Redundancy, premature consensus, noisy debate & Open-ended reasoning, collaborative decision-making \\
Model optimization & Internal reasoning via optimization / latent computation & Stronger amortized reasoning & Training compute and data cost & Underthinking, overthinking, and compute-scaling saturation & High-value domains with reusable reasoning improvements \\
\bottomrule
\end{tabular}
\caption{Cross-dimensional comparison of scaling strategies for LLM reasoning.}
\label{tab:cross_dimension_compare}
\end{table*}

Several recent surveys have covered adjacent areas, including test-time scaling, general LLM reasoning, and post-training scaling~\cite{zhang2025survey,ke2025survey,lai2025survey}. However, these works typically emphasize a particular stage of scaling, a specific reasoning regime, or a broader architectural view of reasoning systems. More generally, existing surveys often organize reasoning methods by technique families (e.g., RAG, CoT, multi-agent systems, and RL). In contrast, our survey focuses on a different question: how different forms of scaling specifically influence reasoning. We organize the literature by \emph{what form of computation or information is scaled} at inference or training time, which enables a unified comparison of otherwise disconnected lines of work in terms of their reasoning gains, costs, limitations, and characteristic failure modes. Specifically, we categorize reasoning scaling into four dimensions. We first discuss \emph{input scaling}, which expands the external information available to the model through larger contexts, retrieval, demonstrations, or memory. We then examine \emph{reasoning step scaling}, which allocates more intermediate computation to decomposition, verification, and search. Next, we study \emph{reasoning round scaling}, in which LLMs iteratively refine their outputs through interaction, such as multi-agent collaboration, debate, and human-in-the-loop feedback. Finally, we consider \emph{training-enabled reasoning}, which improves reasoning by internalizing stronger reasoning behaviors through optimization. Table~\ref{tab:survey_positioning} compares our survey with prior surveys, and Table~\ref{tab:cross_dimension_compare} summarizes the core trade-offs across these four scaling dimensions, including what is scaled, their main benefits and costs, typical failure modes, and the task settings for which they are best suited.

Table~\ref{tab:cross_dimension_compare} summarizes the core trade-offs across these four scaling dimensions, including what is scaled, their main benefits and costs, typical failure modes, and the task settings for which they are best suited. Although these dimensions are often studied separately, they differ systematically in where computation is allocated, what benefits they provide, and what limitations they introduce. The following sections examine each dimension in detail.

By systematically reviewing the scaling of reasoning in LLMs, this survey aims to bridge the gap between empirical scaling strategies and reasoning improvements. Beyond summarizing representative methods, we seek to clarify when and why scaling improves reasoning, where its returns diminish, and what new challenges it introduces. We hope this survey serves as a useful resource for both researchers and practitioners seeking effective, efficient, and reliable ways to advance LLM reasoning.

\begin{figure*}[!t]
\centering
\includegraphics[width=\linewidth]{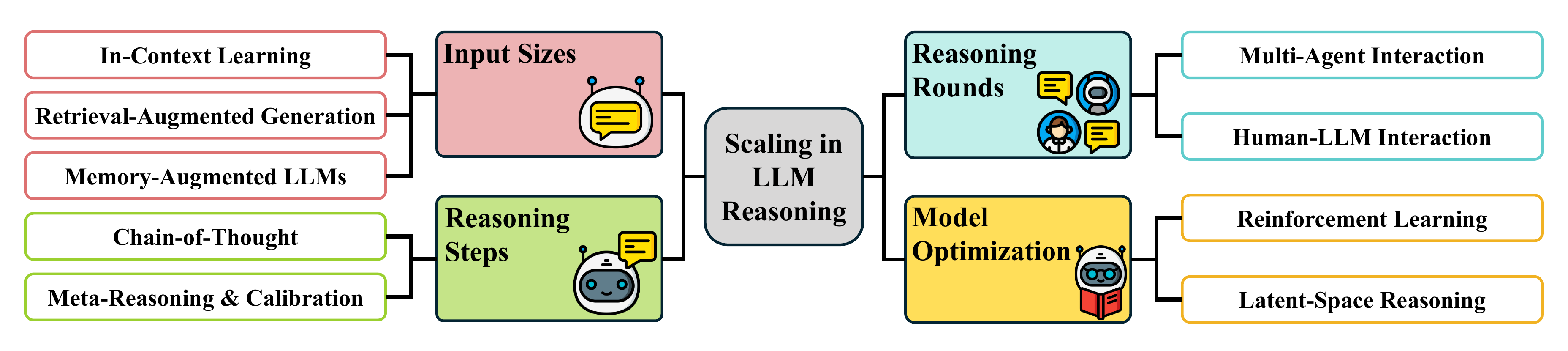}
% \vspace{-0.25in}
\caption{Taxonomy for Scaling in Large Language Model Reasoning.}
\label{fig:framework}
\end{figure*}

% \input{taxmony}

% \vspace{-.05in}
\section{Scaling in Input Sizes}
\begin{figure}
\centering
\includegraphics[width=\linewidth]{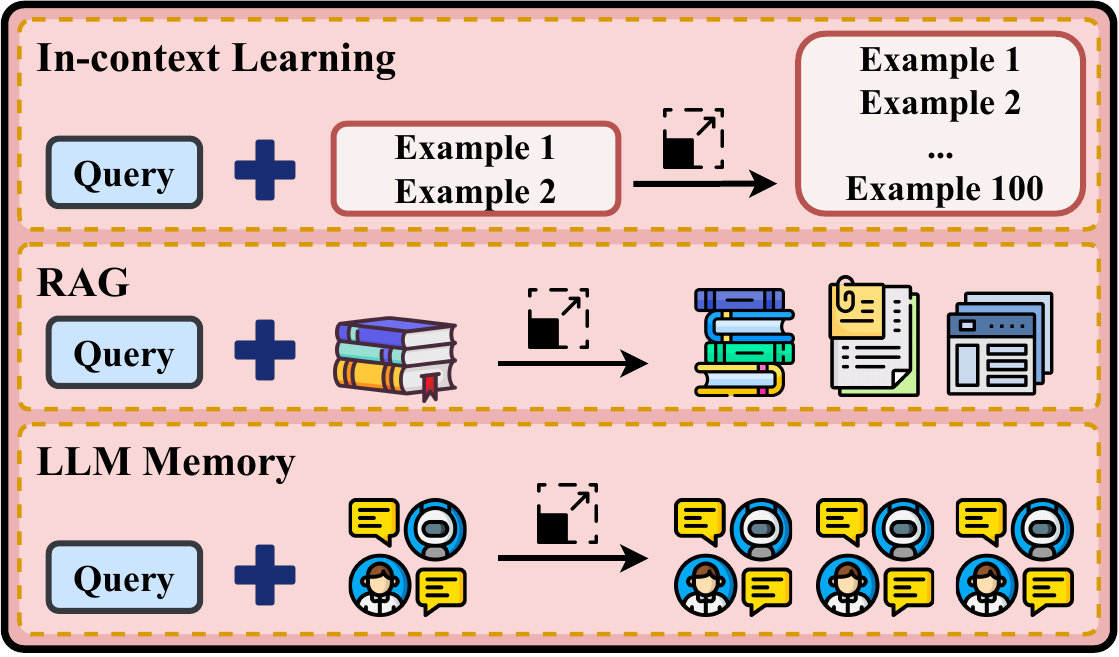}
\caption{Scaling in LLM input sizes.}
\label{fig:input}
\end{figure}
% As LLMs scale, their ability to process larger input contexts becomes increasingly important for enhancing reasoning, retrieval, and adaptability. Providing more contextual information allows models to make more informed and robust inferences. However, longer inputs also bring challenges, including higher computational costs, memory constraints, and efficiency bottlenecks. This section examines key strategies for scaling input sizes—such as ICL, RAG, and memory-augmented LLMs—highlighting their strengths, limitations, and impact on reasoning performance.

Scaling input sizes expands the amount of information available to an LLM during reasoning, rather than increasing the depth of reasoning itself. This dimension improves performance by supplying richer contextual evidence, including demonstrations, retrieved documents, and persistent memory, which can enhance grounding, task adaptation, and long-horizon consistency. Its central trade-off is that additional context often improves coverage and robustness, but also increases retrieval overhead, long-context computation, and the risk of distraction from irrelevant or poorly organized information. In this section, we examine three major strategies for input scaling---ICL, RAG, and memory-augmented LLMs---and analyze how they improve reasoning performance as well as the bottlenecks they introduce.

\subsection{In-Context Learning} 
In-Context Learning (ICL) enables LLMs to adapt to new tasks without parameter updates by conditioning on demonstrations provided in the input prompt. Various algorithms have been developed to improve ICL performance by optimizing demonstration selection~\cite{wang2025mixture,rubin2022learning,ye2023compositional,chen2024fastgas}, ordering~\cite{liu2024let,liu2021makes}, and formatting~\cite{wan2025teach,liu2024context,kim2022self}. More broadly, ICL illustrates a core form of input scaling: increasing the amount of task-relevant context so that the model can better infer the desired behavior from examples alone.
Research has shown that model performance often improves as the number of in-context examples increases~\cite{abedsoltan2024context, min2022rethinking, brown2020language, lu2021fantastically}. However, traditional ICL methods remain constrained by the maximum input context length, which has historically limited them to the few-shot regime~\cite{dong2024survey}. Although some works, such as SAICL~\cite{cai2023scaling}, modify the attention structure to scale ICL to hundreds of demonstrations~\cite{li2023context, li2024paraicl, hao2022structured}, they do not fully explore the broader benefits and challenges of operating with substantially larger demonstration sets.

With the expansion of context windows, researchers have increasingly investigated many-shot ICL, in which models leverage hundreds or even thousands of demonstrations~\cite{bertsch2024context, agarwal2024many}. Scaling from few-shot to many-shot ICL has yielded substantial gains across a wide range of generative and discriminative tasks~\cite{song2024can, zou2024retrieval, park2024iclr}. However, these gains are not unbounded: as the number of in-context demonstrations continues to grow, performance often plateaus and can even decline. This highlights a key limitation of input scaling: more context is only useful when the added information remains relevant, diverse, and well organized. To address these challenges, several methods have been proposed to improve the effectiveness and robustness of many-shot ICL~\cite{baek2024revisiting, zhang2025more, wan2025few}. For example, DrICL~\cite{zhang2025more} adjusts demonstration weights using reinforcement-learning-inspired cumulative advantages to improve generalization, while BRIDGE~\cite{wan2025few} identifies a subset of influential examples and uses them to generate additional high-quality demonstrations.

\vspace{.05in}
\subsection{Retrieval-Augmented Generation} %czh
Retrieval-Augmented Generation (RAG) has become a widely adopted strategy to address the limitations of LLMs, such as hallucinations and restricted generalization to concepts beyond their training data~\cite{jiang2023active,karpukhin2020dense,guu2020retrieval,lewis2020retrieval}. By incorporating retrieved external information, RAG enhances factual grounding and expands the model’s accessible knowledge base. However, traditional RAG operates on short retrieval units, requiring the retriever to scan a massive document corpus to find relevant passages~\cite{qu2020rocketqa,xiong2020approximate,chen2017reading}. This approach is constrained by input context length limitations, making long-context RAG a challenge. A common strategy is document chunking~\cite{sarthi2024raptor,xu2023recomp}, where LLMs retrieve relevant chunks instead of full documents. However, defining optimal chunk boundaries is difficult, often leading to semantic incoherence and contextual loss, which degrade retrieval effectiveness~\cite{li2024uncertaintyrag}. Recent advances in long-context LLMs allow models to process millions of tokens~\cite{team2024gemini}. Integrating RAG with long-context LLMs enables the processing of extended contexts while reducing semantic incoherence in chunked retrieval~\cite{li2024retrieval,xu2024chatqa,xu2023retrieval}.

As input length increases, the burden on retrieval systems grows. LongRAG~\cite{jiang2024longrag} mitigates this by grouping related documents, reducing the number of retrieval operations while maintaining relevance. ReComp~\cite{xu2023recomp} addresses this challenge by compressing retrieved documents into textual summaries before in-context integration, ensuring information remains concise yet informative. 
% Unlimiformer further enhances retrieval efficiency by selecting only the top-k most relevant hidden states, effectively extending retrieval capabilities for long-context reasoning. 
Despite these improvements, a key challenge known as "lost-in-the-middle" bias arises~\cite{liu2024lost}, where LLMs assign less importance to passages in the middle of a retrieved context. MOI~\cite{lee2024inference} counters this bias by aggregating inference calls from permuted retrieval orders, ensuring a more balanced weighting across the retrieved information.

Another dimension of scaling RAG involves expanding the amount of data available at inference time~\cite{wang2023shall,borgeaud2022improving,wang2024instructretro,piktus2021web}. Shao et al.~\cite{shao2024scaling} find that increasing datastore size monotonically improves performance across various language modeling and downstream tasks without clear saturation. Their MASSIVEDS datastore, containing trillions of tokens, is designed to support large-scale retrieval efficiently. Further, Yue et al.~\cite{yue2025inference} explore inference-time scaling, showing that allocating more retrieval computation leads to nearly linear performance gains when optimally distributed. Their work introduces a predictive model for optimizing retrieval parameters under computational constraints.
Together, these findings suggest that input scaling in RAG is effective not only through longer contexts, but also through larger and more searchable external knowledge stores.

\vspace{.05in}
\subsection{Memory-Augmented LLMs} %zt
Scaling reasoning capabilities of LLMs often necessitates extending their effective context beyond the limited token windows supported by existing architectures~\cite{wang2024beyond}. Although increasing context length allows LLMs to process longer sequences, such scaling alone quickly encounters computational bottlenecks and diminishing returns due to quadratic complexity in attention mechanisms~\cite{fu2025sliding}. Moreover, even very long-context models struggle to efficiently capture and retrieve critical historical information from past interactions, leading to degraded reasoning performance over extended contexts~\cite{gao2025u}. To address these limitations, memory augmentation strategies have emerged, enabling LLMs to persistently store, manage, and dynamically retrieve relevant contextual information. Current memory augmentation approaches typically follow two directions: internal architectural modifications to enhance the model’s inherent memory capabilities and external memory mechanisms that extend the model context through additional memory components.

Architectural adaptations focus on internalizing long-term dependencies within the model itself. This includes techniques such as augmenting attention mechanisms to better capture extended context~\cite{liu2023ring,lou2024sparser}, refining key-value cache mechanisms to optimize retrieval efficiency over long sequences~\cite{li2024scbench,liu2025chunkkv}, and modifying positional encodings to enhance length generalization~\cite{zheng2024cape,zheng2024dape}. While effective, these modifications require direct intervention in the model’s structure, making them impractical for proprietary or black-box API-based LLMs.

An alternative approach is the integration of external memory modules to supplement the model’s limited native context window. Summarization-based methods~\cite{lu2023memochat,wang2023recursively,maharana2024evaluating,wang2024adaptive} condense past interactions into structured representations that can be efficiently retrieved during inference. However, fixed-granularity summarization risks fragmenting the discourse, leading to incoherent retrieval. To address this, recent advancements incorporate dynamic memory mechanisms that adaptively refine stored information. RMM~\cite{tan2025prospectretrospectreflectivememory} exemplifies this strategy by leveraging retrospective reflection to improve retrieval selection, ensuring that the model accesses the most relevant and contextually cohesive knowledge.
Scaling memory-augmented LLMs requires balancing efficiency with contextual fidelity. A key challenge is mitigating memory saturation, where excessive storage of past interactions results in retrieval inefficiencies. Techniques such as hierarchical memory organization~\cite{shao2024scaling} and retrieval-conditioned compression~\cite{xu2023recomp} help alleviate this issue by structuring and filtering stored context dynamically. As research progresses, the convergence of retrieval-augmented memory with scalable long-context architectures offers a pro\-mising avenue for enabling LLMs to maintain reasoning consistency over prolonged interactions.

Overall, input scaling improves reasoning primarily by enriching the information available to the model, rather than by increasing the depth of intermediate reasoning or the amount of interactive refinement. This makes it especially effective for knowledge-intensive QA, long-context understanding, and conversational settings that depend on grounding in demonstrations, retrieved evidence, or prior interactions. At the same time, its gains are constrained by context quality, retrieval precision, and the model’s ability to use long inputs effectively. Compared with later dimensions such as reasoning steps and reasoning rounds, the main bottleneck of input scaling is not insufficient inference depth, but whether the added context is relevant, well-structured, and accessible at the right time.

\section{Scaling in Reasoning Steps}
\begin{figure}
\centering
\includegraphics[width=\linewidth]{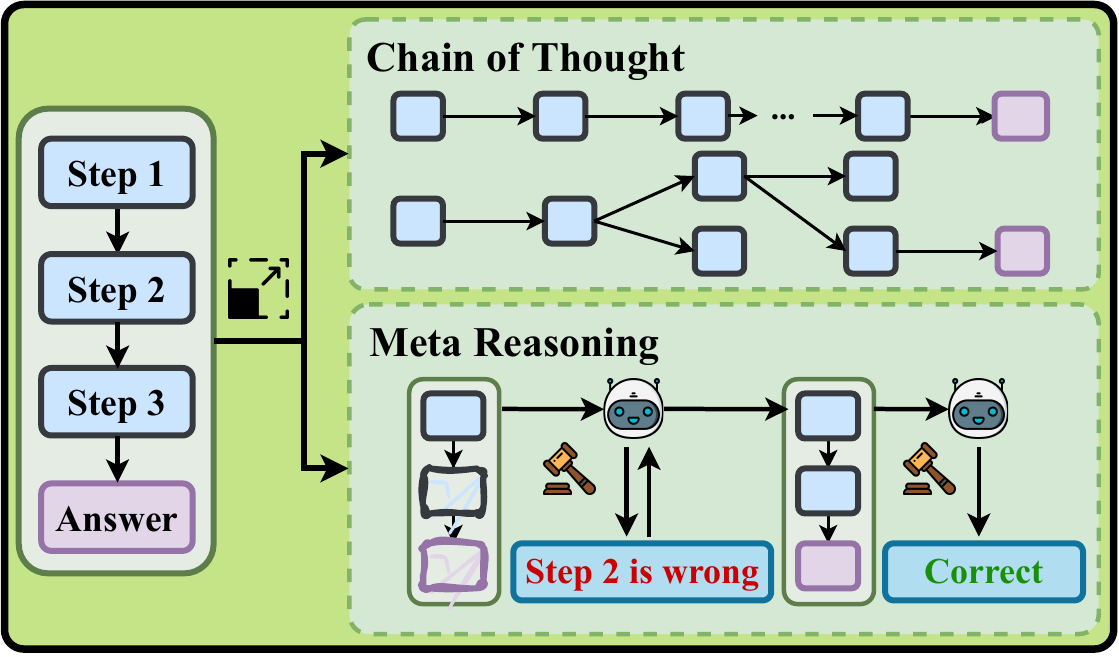}
\caption{Scaling in LLM reasoning steps.}
\label{fig:step}
\end{figure}

% Complex reasoning tasks often require multi-step computation, where models must decompose problems, iteratively refine solutions, and verify correctness. Scaling the depth and breadth of reasoning can enhance logical consistency and problem-solving performance, but it also introduces risks such as overthinking and increased computational cost. This section explores key approaches for scaling reasoning, including Chain-of-Thought prompting and meta-reasoning techniques. We examine methods that improve reasoning by encouraging models to ``think in more steps'' as well as strategies to mitigate the challenges that arise from deeper reasoning processes.

Scaling reasoning steps increases the amount of intermediate computation allocated to solving a problem. Unlike input scaling, which improves reasoning by supplying more contextual information, this dimension improves performance by encouraging models to decompose problems, explore candidate solution paths, iteratively refine intermediate outputs, and verify correctness before committing to an answer. Such additional reasoning depth can substantially improve logical consistency and problem-solving ability, especially on tasks requiring structured multi-step inference. However, it also introduces important trade-offs, including higher token and search costs, longer latency, and the risk that additional reasoning may am: Chain-of-Thought prompting and meta-reasoning techniques, and discuss both their benefits and strategies to mitigate the challenges that arise from deeper reasoning processes.

\subsection{Chain-of-Thought}
% ** Add more content focuses on scaling, potentially remove some nuanced explanations about different ToT/CoT+SC and corresponding variants
% 1. When More is Less: Understanding Chain-of-Thought Length in LLMs: https://arxiv.org/pdf/2502.07266v1
Chain-of-Thought (CoT) prompting has emerged as a key technique for improving the reasoning capabilities of LLMs by eliciting explicit step-by-step deliberation, either through zero-shot prompting~\cite{kojima2022large} or few-shot demonstrations~\cite{wei2022chain}. More broadly, CoT represents a direct form of step scaling: rather than supplying the model with more external information, it allocates more inference-time computation to constructing intermediate reasoning traces. Since LLMs operate probabilistically~\cite{ji2023survey, kumar2025llm}, greedy decoding does not always yield the best reasoning path or final answer~\cite{wang2024chain}. To mitigate this limitation, repeated sampling approaches such as self-consistency~\cite{wang2022self} and Best-of-N~\cite{nakano2021webgpt, brown2024large} generate multiple reasoning chains in parallel and then select the final answer based on criteria such as majority agreement, external reward models, or auxiliary verifiers. They improve robustness by exploring multiple candidate trajectories, while introducing substantial computational overhead.

Although simple parallel sampling is computationally straightforward, it remains inefficient and suboptimal by randomly allocating the test-time computation budget to less promising branches~\cite{wu2024inference, snell2024scaling}. To mitigate this issue, researchers have explored strategies that prioritize promising reasoning paths or intermediate steps over less viable alternatives to effectively prune the search space by applying tree search-enabled reasoning~\cite{wang2022self, yao2023tree, ning2023sot, long2023large, sel2023algorithm, mo2024tree, kim2023tree}
% In particular, tree search-enabled reasoning~\cite{wang2022self, yao2023tree, ning2023sot, long2023large, sel2023algorithm, mo2024tree, kim2023tree} generalizes the CoT approach to further enhance the reasoning capabilities of LLMs.
Generally, it structures the reasoning process as a branching tree, where each node represents a discrete thinking step, and branches correspond to different potential solution paths.
Like CoT which organizes reasoning in a hierarchical manner, tree search-enabled reasoning enables LLMs to decompose intricate problems into manageable components.
However, LLM reasoning with tree search can maintain awareness of multiple hypothesis threads simultaneously and systematically explore the solution space through different search algorithms (e.g., BFS or DFS), making it more powerful for handling complex problems.

The pioneering work CoT-SC~\cite{wang2022self} extends CoT to the tree structure, where multiple CoTs originate from the same initial (root) prompt, forming a ``tree of chains''. The chain that provides the best outcome to the initial question, is selected as the final answer.
Skeleton-of-Thought (SoT)~\cite{ning2023sot} instead effectively harnesses a tree with a specific level of depth.
It performs reasoning through a divide-and-conquer manner, which significantly reduces the generation latency of LLMs.
In the first prompt, the LLM is instructed to generate a skeleton of the answer, i.e., a list of points that can be answered independently.
For each point, a new prompt is issued in parallel to address only the corresponding part of the question.

Recently, numerous studies have explored Tree of Thoughts (ToT)~\cite{yao2023tree, long2023large} for tree search-enabled reasoning.
Compared to CoT-SC where multiple CoTs originate from the same initial (root) prompt, ToT employs a tree structure to decompose a problem into subproblems and solve them using separate LLM prompts.
Unlike ToT using multiple prompts, Algorithm of Thoughts (AoT)~\cite{sel2023algorithm} uses only a single prompt with in-context examples formulated in an algorithmic fashion.
Tree of Uncertain Thought (TouT)~\cite{mo2024tree} enhances ToT with local uncertainty scores by incorporating the variance of multiple LLM responses into the state evaluation function.
Tree of Clarifications (ToC)~\cite{kim2023tree} focuses on answering ambiguous questions using ToT.
It first retrieves relevant external information and then recursively prompts an LLM to construct a disambiguation tree for the question.

% TODO: add MCTS and S1 (maybe also relevant works for sequential cot)

\vspace{.05in}
\subsection{Meta-Reasoning and Calibration} %xb
Numerous works~\cite{dhuliawala2024chain, paul2024refiner, gou2023critic, zelikman2024quiet, jiang2023flare} have shown that LLMs have inherited capabilities of self-correction with proper prompt engineering.
Typically, an LLM can self-reflect its responses by generating feedback on its answers.
It first generates an initial response to an input question.
Next, it generates feedback given the original input and its initial response.
Finally, it generates a refined response given the input, initial response, and feedback.
Generally, self-correction may rely on different sources of feedback, including intrinsic prompts and external information.
Intrinsic prompts let LLMs generate feedback on their own responses.
For example, CoVe~\cite{dhuliawala2024chain} plans verification questions to check an initial response and then systematically answers those questions in order to finally produce an improved revised response.
FLARE~\cite{jiang2023flare} performs self-correction by iteratively generating a temporary next sentence and check whether it contains low-probability tokens.
In contrast, external information enables LLMs to rely on external tools, such as external knowledge from search engines, oracle information, and task-specific metrics, to enhance self-correction.
For example, REFINER~\cite{paul2024refiner} interacts with a critic model that provides automated feedback on the reasoning.
CRITIC~\cite{gou2023critic} interacts with external tools like search engines and code interpreters to verify the desired aspects of an initial output and subsequently amends the output based on the critiques from the verification.

One major concern centers around the efficiency of self-refinement: LLMs need to generate feedback and refined responses iteratively, which can significantly increase the inference time of LLMs.
To overcome the scaling issue, Quiet-STaR~\cite{zelikman2024quiet} designs a tokenwise parallel
sampling algorithm, using learnable tokens indicating a thought’s start and end, and an extended teacher-forcing technique.
Another concern is caused by generation-time correction.
Prevalent self-correction approaches are based on generation-time correction, heavily depending on the capacity of the critic model to provide accurate quantifiable feedback for intermediate outputs.
Nevertheless, this might be quite challenging for many NLP tasks with long token sizes, such as summarization--- the summary can be accurately assessed only after the entire summary is generated.
This limitation makes generation-time correction infeasible in many NLP tasks.
One solution to this issue is post-hoc correction~\cite{pan2024automatically}. Unlike general generation-time correction which generates feedback on the intermediate reasoning steps, post-hoc correction involves refining the output after it has been generated.

Overall, step scaling improves reasoning by allocating more computation to decomposition, search, verification, and correction. Compared with input scaling, which primarily addresses deficiencies in available evidence or context, step scaling is most effective when the main bottleneck lies in the reasoning process itself, as in mathematical, logical, and planning tasks. However, its gains are constrained by search efficiency, verification reliability, and the model’s ability to avoid overthinking or compounding early mistakes. Relative to later dimensions such as reasoning rounds, which broaden reasoning through interaction, step scaling deepens a single model’s inference process and therefore offers a more direct but often less diverse form of reasoning improvement.

% However, a few studies~\cite{gou2023critic, huang2023large} report that self-correction may not always improve LLM performance in some tasks.

% Automatically correcting large language models: Surveying the landscape of diverse self-correction strategies
% Language agent tree search unifies reasoning acting and planning in language models
% Quiet-star: Language models can teach themselves to think before speaking
% Livecodebench: Holistic and contamination free evaluation of large language models for code
% Fincon: A synthesized llm multi-agent system with conceptual verbal reinforcement for enhanced financial decision making
% When Can LLMs Actually Correct Their Own Mistakes? A Critical Survey of Self-Correction of LLMs
% LLMs cannot find reasoning errors, but can correct them given the error location
% Cumulative reasoning with large language models
% Cycle: Learning to self-refine the code generation
% Self-reflection in llm agents: Effects on problem-solving performance

\section{Scaling in Reasoning Rounds}
\begin{figure}
\centering
\includegraphics[width=\linewidth]{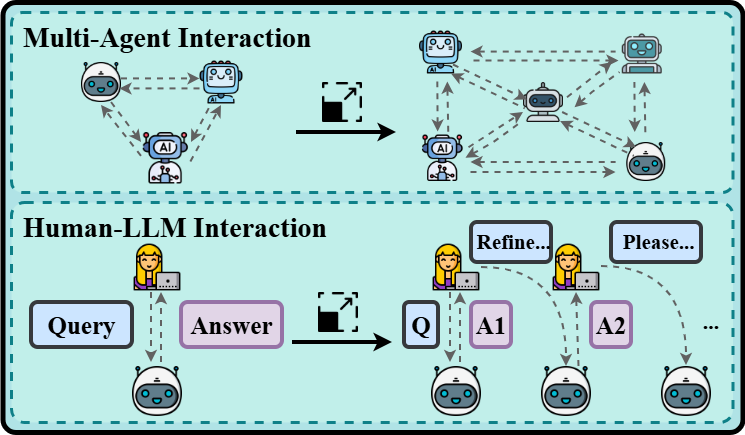}
\caption{Scaling in reasoning rounds.}
\label{fig:round}
\end{figure}
% Beyond single-step or sequential reasoning, iterative multi-round reasoning enables LLMs to refine responses, debate alternatives, and integrate external feedback. However, scaling the number of reasoning rounds introduces challenges related to efficiency, redundancy, and diminishing performance returns. This section explores key approaches that leverage iterative interaction, including multi-agent collaboration, debate-based reasoning, and human-LLM interaction.

% Beyond single-step or purely sequential inference, many reasoning processes benefit from iterative multi-round interaction, where models refine their answers, incorporate feedback, and explore alternative solution paths. Increasing the number of reasoning rounds, however, raises important considerations: additional iterations can improve reasoning depth but also introduce computational overhead, redundancy, and diminishing returns. In this section, we examine how scaling interaction rounds shapes reasoning across two major paradigms—multi-agent interaction, where multiple LLMs coordinate or specialize across rounds, and human–LLM interaction, where iterative human feedback guides and stabilizes the model's reasoning.

Scaling reasoning rounds expands the reasoning process thr\-ough iterative interaction rather than through a single forward pass or a single model’s internal chain of thought. Unlike input scaling, which enriches the information available to the model, and step scaling, which deepens intermediate computation within one reasoning trajectory, round scaling improves performance by introducing repeated communication, critique, and refinement across multiple turns. This additional interaction can increase diversity of perspectives, expose hidden errors, and support iterative improvement, but it also comes with significant trade-offs, including coordination overhead, latency, redundancy, and diminishing returns as the number of rounds grows. This section examines two major paradigms of round scaling: multi-agent interaction, where multiple LLMs coordinate or debate across rounds, and human--LLM interaction, where iterative human feedback guides and stabilizes the reasoning process.

\subsection{Multi-Agent Interaction}
Multi-agent interaction has emerged as a powerful paradigm for scaling LLM reasoning by enabling multiple models to iteratively exchange information, challenge assumptions, and refine outputs. Broadly, existing approaches fall into two major categories: \emph{collaborative} frameworks, which emphasize specialization and division of labor, and \emph{debate-based} frameworks, which introduce adversarial reasoning to expose errors and strengthen the final decision.

In collaborative settings, multiple LLMs work together in a coordinated manner to achieve improved problem-solving capabilities~\cite{liang2023encouraging,
kenton2024scalable,li2023camel}. In particular, in these frameworks, each LLM (agent) is assigned a distinct role—such as planner, executor, verifier, or critic—and iteratively refines its output through structured interactions with other agents~\cite{xu2024languageagentsreinforcementlearning}. For example, CAMEL~\cite{li2023camel} introduced a framework where LLM agents assume different personas and interact through structured role-playing, enabling more effective task completion through multi-turn communication. 
The core idea is to enhance the specialization and division of labor among LLMs, ensuring that different agents contribute unique perspectives to improve overall task performance. Unlike single-agent systems, which rely on an LLM’s internal reasoning capability~\cite{guo2024large, weng2023agent}, multi-agent frameworks distribute tasks across multiple agents that engage in iterative interactions~\cite{li2023camel}.

Increasing the number of agents can improve task diversity and allow for role specialization, where different agents assume distinct functions such as problem decomposition, tool usage, or evaluation~\cite{guo2024largelanguagemodelbased}. Research has demonstrated that larger multi-agent systems can achieve greater accuracy and better adaptability in open-ended reasoning tasks, as seen in software development frameworks like MetaGPT~\cite{hong2023metagptmetaprogrammingmultiagent}. However, these gains are not monotonic. Beyond a certain scale, performance may plateau or even decline due to conflicting reasoning paths, redundancy, and growing coordination overhead~\cite{liang2023encouraging}. 
Similarly,~\cite{rasal2024llm} shows that structured dialogue among LLM agents improves reasoning depth and solution diversity, but also finds that too many interaction rounds lead to diminishing returns, as agents increasingly reinforce each other’s biases rather than contribute genuinely new insights. These findings suggest that naive scaling of agent count or communication depth is insufficient; effective round scaling requires careful coordination protocols and complementary role assignment.
%
% Nevertheless, introducing hierarchical structures, where some LLMs serve as supervisors while others act as task executors, has shown consistent improvements in task accuracy and efficiency~\cite{cai2023large}.
Several works therefore introduce explicit communication structures to mitigate these issues. Hierarchical frameworks, in which some LLMs act as supervisors while others serve as task executors, have shown consistent gains in both accuracy and efficiency~\cite{cai2023large}. 
%
%While increasing the number of agents diversifies perspectives, it can also amplify existing biases within the system. Prior research has shown that LLMs exhibit societal biases, such as those related to gender, race, and political ideology\cite{wan2023kelly, dong2024disclosure}. In a multi-agent setting, where agents reinforce each other’s outputs through iterative interactions, pre-existing biases in individual agents can be magnified within the collective decision-making process\cite{zhou2024empirical, kotek2023gender}.
%
Another interesting finding is introduced in LLM Harmony~\cite{rasal2024llm}, which optimizes inter-agent communication by structuring dialogue between multiple LLM agents. %
Instead of simple turn-based exchanges, this framework enables agents to dynamically negotiate task objectives, delegate subtasks, and refine outputs iteratively. 
The results suggest that scaling the number of interacting agents improves performance only when they are given complementary roles, while increasing homogeneous agents leads to redundant reasoning patterns. 

In contrast to collaborative frameworks, debate-based methods deliberately assign \emph{adversarial} roles to LLMs and often introduce an explicit judge. In these setups, each agent acts as a debater that challenges others and attempts to persuade a judge, with the goal of surfacing errors and stronger arguments rather than directly solving subtasks. A pioneering example is Multi-Agent Debate (MAD)~\cite{liang2024encouraging}, which proposes a structured debate protocol with a ``tit-for-tat'' mechanism: multiple debaters exchange arguments over several rounds, and a designated judge aggregates the discussion to reach a final decision. The key idea is to amplify disagreement and critical scrutiny. Compared with self-reflection approaches~\cite{madaan2023self,shinn2023reflexion}, MAD induces stronger disagreement, helping to avoid premature convergence on incorrect answers. Building on this idea, subsequent debate-based frameworks improve reasoning robustness and factual accuracy by refining debate protocols and judge designs~\cite{du2023improving}.
% \subsection{Debate-Based Reasoning}%sw
% Beyond the general framework of leveraging multiple LLMs for collaborative task execution, researchers have also explored the use of LLMs in multi-round reasoning to enhance reasoning effectiveness. Specifically, in these frameworks, each LLM (or agent) functions as a debater, engaging in discourse to challenge and persuade others while refining its own reasoning through iterative exchanges. A pioneering work in this area, Multi-Agent Debate (MAD)~\cite{liang2024encouraging}, introduces a framework in which multiple agents engage in a structured debate following a "tit-for-tat" mechanism, with a designated judge overseeing the discussion to arrive at a definitive answer. The core idea is to encourage diverse perspectives among agents, fostering deeper contemplation and critical thinking. The authors demonstrate that the debate framework leads to significantly higher disagreement levels compared to Self-Reflection~\cite{madaan2023self,shinn2023reflexion}, thereby reducing the risk of models converging on incorrect answers. Given these advantages, researchers have proposed various debate-based frameworks that enhance both reasoning capabilities and factual accuracy~\cite{du2023improving}.
%

The scaling effect in debate frameworks manifests in multiple dimensions. In \cite{khan2024debating}, the authors find that when employing a judge LLM to evaluate responses from debater LLMs, increasing the number of debate rounds does not necessarily lead to greater clarity—especially for weaker models, where additional rounds introduce confusion rather than improving accuracy. However, in consultancy-based interactions, where a single LLM attempts to persuade a judge LLM, the judge's accuracy improves over successive rounds. Notably, enhancing the persuasiveness of debater LLMs—making them more effective at convincing the judge—has been shown to yield performance improvements. This scaling effect provides further insights into optimizing debate-based reasoning frameworks. Similarly, \cite{michael2023debate} suggests that scaling LLM debates with increasingly skilled debaters (e.g., progressing from AI to human debaters) enhances oversight mechanisms, improving overall debate efficacy, whereas consultancy frameworks tend to perform worse under similar conditions.
Distinct from these approaches, CIPHER~\cite{pham2024let} proposes embedding-based communication to facilitate debate, enabling smaller LLMs to retain stronger debate capabilities by mitigating information loss. Their findings indicate that increasing the number of debate rounds improves performance up to a threshold of three rounds, beyond which additional rounds provide diminishing returns.
Overall, multi-agent interaction shows that round scaling can improve reasoning not only by increasing deliberation length, but also by introducing complementary roles, disagreement, and iterative critique. At the same time, it reveals a defining challenge of this dimension: additional rounds are helpful only when they generate genuinely new information or perspectives, rather than repeated, biased, or poorly exchanges.

\vspace{.05in}
\subsection{Human-LLM Interaction} %zt
% Scaling LLM reasoning is not solely a function of model size and context window but also hinges on the quality and depth of human interactions~\cite{alsagheer2024comparing}. Human-in-the-loop frameworks~\cite{wu2022survey} enhance LLM performance by integrating iterative refinement, feedback-driven prompting, and adaptive response generation. This interaction paradigm shifts LLMs from static inference engines to dynamically evolving agents capable of learning from user interventions.

Reasoning rounds can also be scaled through iterative interaction between humans and LLMs. In this setting, improvement does not come from communication among multiple models, but from repeated user guidance that helps the model clarify goals, correct mistakes, and refine its responses. This human-in-the-loop paradigm shifts LLMs from static inference engines to adaptive assistants whose reasoning can be steered and stabilized through feedback~\cite{alsagheer2024comparing,wu2022survey}.

Recent work explores multi-turn reasoning scenarios where users provide incremental clarifications or corrections, allowing models to refine their responses iteratively~\cite{madaan2023self,krishna2024understanding}. This process mirrors how humans engage in collaborative problem-solving, gradually converging on an accurate and well-structured answer. Methods such as self-reflection prompting~\cite{shinn2023reflexion} and feedback-based reinforcement learning~\cite{chen2025reinforcement} demonstrate improvements in factual consistency and reasoning depth by enabling LLMs to assess and revise their outputs.

A key challenge in human-LLM interaction is balancing efficiency with adaptability. Over-reliance on explicit feedback mechanisms can introduce cognitive overhead for users, while insufficient adaptability limits the model’s ability to incorporate nuanced human guidance. Recent strategies mitigate this tradeoff through adaptive interaction mechanisms, such as retrieval-enhanced dialogue memory~\cite{pan2025memory} and user-intent modeling~\cite{li2023web}, allowing LLMs to anticipate user needs and refine responses proactively.

As interaction frameworks scale, ensuring alignment with human cognitive processes remains critical. Fine-tuning strategies that incorporate user feedback loops have shown promise in enhancing model interpretability and trustworthiness~\cite{kim2025human}. Furthermore, inference-time intervention mechanisms~\cite{manvi2024adaptive,tan2024tuning} enable LLMs to allocate computational resources efficiently based on user engagement patterns. By refining the synergy between LLMs and human oversight, interactive reasoning systems hold the potential to scale beyond static prompt-response architectures, evolving towards more adaptive and contextually aware AI assistants.

Overall, round scaling improves reasoning by broadening the inference process through interaction, critique, and iterative refinement. Compared with step scaling, which deepens a single reasoning trajectory, round scaling introduces external feedback and perspective diversity, making it particularly useful for open-ended reasoning, oversight, and collaborative decision-making. However, its gains are constrained by communication quality, role complementarity, and coordination efficiency, and its characteristic failure modes include redundancy, premature consensus, noisy debate, and excessive reliance on user effort. Relative to training-enabled reasoning, which seeks to internalize better reasoning policies in advance, round scaling remains more flexible at inference time but also more dependent on well-designed interaction protocols.

% \section{Scaling in Training-enabled Reasoning}
\section{Scaling in Model Optimization}
\begin{figure}
\centering
\includegraphics[width=\linewidth]{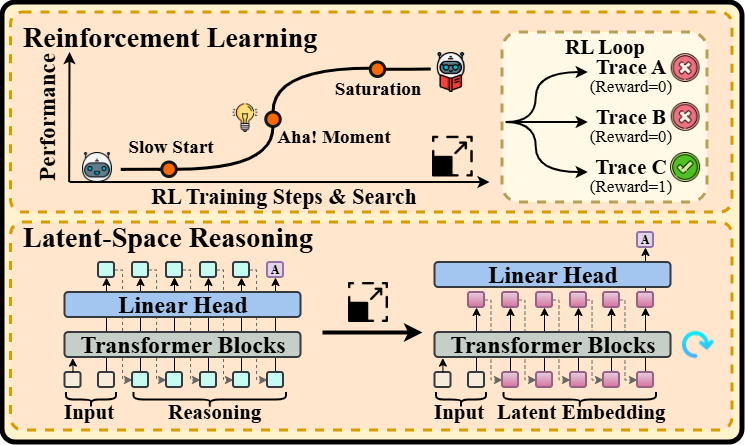}
\caption{Scaling in model optimization.}
\label{fig:optm}
\end{figure}
% Beyond inference-time techniques, scaling model optimization provides another pathway to enhance LLM reasoning by directly expanding the model’s effective computation through training dynamics and latent-space processing. Reinforcement learning (RL)–based methods scale reasoning by optimizing the model’s policies over increasingly complex tasks, aligning behavior with human intentions and enabling deeper multi-step inference. Complementing RL approaches, latent-space reasoning methods scale internal computation without increasing sequence length or model size. By iterating over hidden representations, such as in-looped transformer, these methods allow models to perform additional internal reasoning steps, effectively scaling “thinking time” while keeping inference cost manageable. Together, Reinforcement Fine-Tuning (RFT) and latent-space reasoning illustrate how optimization-time scaling can deepen LLM reasoning capacity, offering alternatives to traditional scaling via larger models or longer explicit reasoning traces.

Scaling model optimization improves reasoning not by supplying more external information or extending explicit inference trajectories, but by strengthening the model’s internal reasoning capacity through training or latent computation. Unlike the previous three dimensions, which primarily allocate additional resources at inference time, this dimension seeks to internalize better reasoning policies in advance or to expand internal computation without proportionally increasing explicit reasoning length. Reinforcement learning (RL)–based methods scale reasoning by optimizing the model’s policies over increasingly complex tasks, aligning behavior with human intentions and enabling deeper multi-step inference. Complementing RL approaches, latent-space reasoning methods scale internal computation without increasing sequence length or model size. By iterating over hidden representations, such as an in-looped transformer, these methods allow models to perform additional internal reasoning steps, effectively scaling “thinking time” while keeping inference cost manageable. Together, Reinforcement Fine-Tuning (RFT) and latent-space reasoning illustrate how opt\-imization-time scaling can deepen LLM reasoning capacity, offering alternatives to traditional scaling via larger models or longer explicit reasoning traces.

% \subsection{Reinforcement Learning for Fine-Tuning}
\subsection{Reinforcement Learning}
% **This subsection focus on**: scaling training steps in RL, scaling length of response (reasoning chain) in RL and scaling data samples during RL. Also some limitation on whether scaling is useful.

% ** Add more scaling related content:
% ** General Investigation on scaling RL
% 1. Kimi k1.5: Scaling Reinforcement Learning with LLMs: https://arxiv.org/abs/2501.12599
% 2. Advancing Language Model Reasoning through Reinforcement Learning and Inference Scaling: https://arxiv.org/abs/2501.11651
% 3. DAPO: An Open-Source LLM Reinforcement Learning System at Scale: https://arxiv.org/abs/2503.14476
% 4. DeepSeek-R1: Incentivizing Reasoning Capability in LLMs via Reinforcement Learning: https://arxiv.org/abs/2501.12948
% ** Replication of o1/R1 style RL training
% 1. Logic-RL: Unleashing LLM Reasoning with Rule-Based Reinforcement Learning
% 2. Training Reasoning Model with Dynamic Advantage Estimation on Reinforcement Learning: https://five-stetson-b51.notion.site/Training-Reasoning-Model-with-Dynamic-Advantage-Estimation-on-Reinforcement-Learning-1a830cc0904681fa9df3e076b6557a3e (ADOPA: A Scalable Paradigm for Steering Learning Trajectories)
% 3. 

Although previous studies have shown that distilling knowledge from superior LLMs, regardless of whether supervised fine-tuning (SFT) data are amassed in large quantities or carefully curated~\cite{zhou2023lima, ye2025limo}, can enhance the reasoning abilities of smaller models for solving complex tasks~\cite{shridhar2023distilling, magister2022teaching, ho2022large}, recent studies contend that, merely increasing the volume of SFT data typically yields only a log-linear performance improvement~\cite{yuan2023scaling}. Moreover, models trained exclusively on SFT data tend to overfit by memorizing the training set, thereby struggling to generalize to out-of-distribution (OOD) tasks~\cite{chu2025sft}. To mitigate these issues, reinforcement learning (RL) has emerged as a key approach in LLM post-training, aligning models with human preferences~\cite{ouyang2022training, rafailov2023direct} and enhancing their reasoning abilities~\cite{shao2024deepseekmath, yang2024qwen2, guo2025deepseek}.

% Fine-tuning LLMs using RL involves optimizing the model, typically via policy gradient algorithms such as Proximal Policy Optimization (PPO)~\cite{schulman2017proximal}, to maximize the response's reward. This process can leverage explicit reward models such as outcome reward models (ORM), which compute reward based on the entire response or using heuristic or rule-based functions to assess the final answer, and process reward models (PRM), which compute reward at each intermediate step, either from human annotations~\cite{uesato2022solving, lightman2023let} or Monte Carlo (MC) estimation~\cite{wang2023math, zhang2025lessons}.

% A key challenge in PPO is its computational overhead~\cite{ahmadian2024back}. Since PPO constrains policy updates to remain close to a reference model, it requires an actor, a reference, and a reward model when computing reward, and further needs a critic model to estimate the advantage using Generalized Advantage Estimation (GAE)~\cite{schulman2015high}. To mitigate this issue and stabilize the training process, ~\citet{ahmadian2024back} and ~\citet{hu2025reinforce++} suggest replacing the complicated PPO with vanilla REINFORCE by modeling the entire generation as a single action and removing the critic model in PPO. ~\citet{shao2024deepseekmath} introduces GRPO, which substitutes GAE in PPO with moving average of all rewards from the group of responses of the same prompt. These simplified PPO variants enhance scalability, making large-scale training more practical. 

Recent studies indicate that conducting RL-based fine-tuning following SFT can further improve the reasoning abilities of LLMs. ReFT~\cite{luong2024reft} first performs a warm-up SFT on distilled CoT data followed by fine-tuning the SFT model using Proximal Policy Optimization (PPO)~\cite{schulman2017proximal} on the same training questions, which eventually leads to significant performance gains on mathematical reasoning tasks.
T1~\cite{hou2025advancing} employs a similar training strategy, but emphasizes scaling sampling diversity during RL training through techniques such as high-temperature sampling, on-policy KL normalization, and rule-based reward penalties for undesirable repetition responses. They observed that increasing the number of sampled responses, raising the sampling temperature during RL training, and extending inference-time reasoning steps collectively contribute to improved reasoning performance.
DeepSeek-R1~\cite{guo2025deepseek} shares a similar strategy as ReFT but employs self-training by directly applying Group Relative Policy Optimization (GRPO)~\cite{shao2024deepseekmath} to the base model. This base model is then used to generate long-form CoT data for the warm-up SFT stage, after which GRPO is applied again to the SFT model, ultimately achieving reasoning performance comparable to OpenAI-o1~\cite{jaech2024openai}. They observed an ``aha-moment'' during the training of DeepSeek-R1-Zero, where the model learned to rethink as the response length increased.
% Kimi k1.5~\cite{team2025kimi}
Following DeepSeek-R1, recent works observed similar ``aha-moment'' and think related words on different tasks, including real-world software engineering~\cite{wei2025swe}, logical puzzles~\cite{xie2025logic}, and automated theorem proving~\cite{dong2025stp} when scaling up the training steps and response length using RL-based fine-tuning.

% However, reasoning models trained with RL to generate long CoT responses may also encounter challenges such as ``underthinking''~\cite{wang2025thoughts}, where models frequently switch between reasoning branches without engaging in deep thought, and ``overthinking''\cite{chen2024not}, which suggests that excessive reasoning on simple questions can sometimes degrade performance. Additionally, previous studies~\cite{hou2024does} argue that scaling the number of response samples and increasing the size of the policy model, while keeping the reward model fixed, is less efficient compared to scaling during pre-training.

At the same time, RL-trained reasoning models that produce long CoT traces can exhibit notable failure modes, including ``underthinking''~\cite{wang2025thoughts}, where the model frequently switches between shallow reasoning branches without engaging in sustained deliberation, and ``overthinking''~\cite{chen2024not}, where excessive reasoning on simple instances can degrade accuracy. Beyond these empirical observations, recent work has begun to systematically characterize the compute–scaling behavior of RL post-training. ScaleRL~\cite{khatri2025art} demonstrates that RL reward improvements follow a predictable sigmoidal compute–performance curve, with early low-gain regions, a sharp transition phase, and a clear asymptotic limit. Their analysis shows that many commonly tuned components, including curriculum design, normalization, and loss aggregation, primarily affect compute efficiency, whereas only a small subset of architectural or algorithmic choices (e.g., loss formulation, precision handling, off-policy setup) materially shifts the ultimate performance ceiling. In parallel, ProRL~\cite{liu2025prorl} provides complementary evidence that prolonged RL optimization can elicit qualitatively new reasoning strategies even in relatively small models, though it does not explicitly analyze the predictability of compute scaling. Taken together, these studies suggest that RL post-training is not only empirically effective but also exhibits a structured, saturating scaling pattern, underscoring the importance of understanding compute–performance curves rather than evaluating methods solely at isolated endpoints.

% Add challenges on scaling the questions to open-domain. Scaling can also be put on the verifier side, from traditional RLAIF to rule-based reward design to critic rm and recent generative rm used by deepseek.

% These challenges highlight the need for more effective scaling strategies and a balanced approach to reasoning in RL-trained models to optimize both efficiency and performance.
% Efficient long2short CoT by Kimi k1.5 by applyinv length penalty

\subsection{Latent-Space Reasoning}

A second pathway for optimization-based scaling increases reasoning capacity by allocating additional computation in latent space rather than through longer explicit reasoning traces. In explicit reasoning~\cite{wei2022chain}, models generate intermediate natural-language steps before producing a final output. While such reasoning improves interpretability and decomposition, it can also be verbose and computationally expensive. Latent-space reasoning aims to address this limitation by performing additional computation over hidden representations without requiring every intermediate thought to be verbalized~\cite{deng2023implicit, shen2025efficient}. This makes it possible to scale ``thinking time'' without proportionally increasing sequence length or model size.

Several recent methods instantiate this idea in different ways. Deng et al.~\cite{deng2023implicit} propose distilling multi-step reasoning into latent representations across layers, allowing the model to solve complex problems in a single forward pass while improving efficiency and scalability. CoCoMix~\cite{tack2025llm} trains LLMs to predict selected semantic concepts directly from hidden states; by interleaving token embeddings with high-level continuous concepts, it enhances abstract reasoning while reducing data and computation costs. More broadly, this line of work reflects a key observation: natural language is not always the most efficient substrate for reasoning. Hao et al.~\cite{hao2024training} argue that many word tokens mainly serve textual coherence rather than reasoning itself, whereas only certain critical tokens require deeper planning. Based on this insight, Coconut~\cite{hao2024training} iteratively processes hidden states and enables parallel exploration of multiple reasoning paths in latent space. Additional work explores how iterative latent computation can deepen reasoning without requiring parameter expansion. ITT~\cite{chen2025inner}, for example, dynamically allocates computation to critical tokens and iteratively refines hidden representations. The same iterative paradigm has also been explored for test-time scaling~\cite{geiping2025scaling, mohtashami2023cotformer}, where repeated latent transformations improve efficiency relative to explicitly lengthening verbalized reasoning. Similarly, Saunshi et al.~\cite{saunshi2025reasoning} show that model depth can effectively be scaled under a limited parameter budget through looping, introducing a new scaling paradigm based on iterative latent-space transformations rather than simply enlarging the model. Collectively, these methods suggest that deeper reasoning need not always correspond to longer visible chains of thought; in some settings, the key resource being scaled is internal computation itself.

Overall, optimization-based scaling improves reasoning by internalizing stronger reasoning policies or by expanding internal computation through latent iterative processing. Compared with input scaling, step scaling, and round scaling, which primarily invest additional resources at inference time, this dimension shifts the trade-off toward up-front optimization cost in exchange for potentially reusable reasoning improvements across many downstream tasks. This makes it particularly attractive in high-value settings where stronger reasoning behavior can be amortized over repeated deployment. However, its gains are constrained by optimization stability, reward fidelity, transferability, and the difficulty of understanding how internalized or latent reasoning generalizes beyond the training conditions.

\section{Application} 
\subsection{AI Research}
Scaling in LLMs has fundamentally reshaped AI research, both extending traditional domains and opening entirely new research avenues. This section explores how scaling has influenced three critical areas: LLM-as-a-Judge, fact-checking, and dialogue systems.

\vspace{.05in}
\noindent\textbf{LLM-as-a-Judge.}
Using LLMs to evaluate model outputs or other models has emerged as a pivotal research direction, enabling evaluation at scale beyond traditional approaches and human assessment~\cite{li2024generation}. Notably, larger models demonstrate a significantly higher correlation with human preferences compared to their smaller counterparts~\cite{zheng2023judging}.
To further improve evaluation quality, recent work has explored multi-step reasoning processes~\cite {saha2025learning}, where scaling the number of reasoning steps enhances evaluation capabilities~\cite{chiang2025tract}. Additionally, scaling across multiple judge models has emerged as an effective approach to improve evaluation reliability~\cite{liang2024debatrix}. Different LLMs functioning as agents collaborate through multi-round discussions before reaching a final judgment, thereby enhancing evaluation consistency~\cite{qian2025enhancing}.

\vspace{.05in}
\noindent\textbf{Fact-Checking.}
The capacity of AI systems to generate misinformation has driven substantial research into automated fact checking~\cite{wolfe2024impact, demartini2020human, zhou2023synthetic}.
Initial fact verification approaches relied on smaller models with limited contextual understanding, primarily focusing on matching claims to evidence~\cite{demartini2020human}. Large-scale LLMs have shown remarkable fact-checking capabilities by supporting fact-checkers with their extensive knowledge and sophisticated reasoning~\cite{tang2024minicheck}.
Scaling in reasoning steps has been demonstrated to improve claim detection, making the process more methodical~\cite{sawinski2023openfact}.
Additionally, RAG has been employed for evidence-backed fact-checking with reduced hallucination and improved performance, with performance scaling with the number of retrieved documents~\cite{singhal2024evidence}.
Multi-agent systems have been widely implemented for fact-checking, where multiple imperfect fact-checkers can collectively provide reliable assessments~\cite{verma2025multi}.

\vspace{.05in}
\noindent\textbf{Dialogue Systems.}
Dialogue systems represent the most visible application of LLM scaling~\cite{yi2024survey, zheng2023lmsys, friedman2023leveraging}, where advances in context length, reasoning steps, and training data have dramatically transformed interactive capabilities.
Enhanced context handling has significantly impacted dialogue coherence and consistency. Scaling of context provides dialogue agents with more information, enabling more informative long-term conversations~\cite{bang2015example,tan2025prospectretrospectreflectivememory}.
External augmentation has been widely adopted to facilitate long-term dialogue as well. Commonly integrated external knowledge, including commonsense~\cite{wang2024unims}, medical~\cite{chen2023llm}, and psychological~\cite{chen2023soulchat} knowledge, serves as
supplementary guidance for the reasoning process,
ensuring logical coherence across extended contexts.
Multi-agent dialogue systems have also demonstrated exceptional capabilities, where multiple LLMs collaborate to comprehensively evaluate and select the most appropriate responses~\cite{fang2024multi}.

\subsection{Production}
The scaling reasoning capabilities of LLMs have significantly enhanced production applications, particularly in software development, data science workflows, and interactive AI systems. This subsection discusses these areas with illustrative examples.

\vspace{.05in}
\noindent\textbf{Software Development.}
The scaling reasoning capabilities of LLMs enhance software development by enabling a better understanding of complex coding tasks and facilitating accurate multi-step reasoning over intricate software dependencies and structures. Advanced reasoning techniques, such as chain-of-thought prompting, allow code-generation assistants to systematically approach and solve coding tasks~\cite{chen2021evaluating,jiang2024survey}. Furthermore, structured reasoning strategies can effectively handle larger coding contexts and reduce developer cognitive load during debugging and iterative improvement processes~\cite{jiang2024survey}.

\vspace{.05in}
\noindent\textbf{Data Science Workflows.}
Scaling reasoning in LLMs substantially improves data science workflows by enabling sophisticated analytical and exploratory tasks. Multi-step reasoning allows LLMs to iteratively explore, interpret, and synthesize insights from diverse datasets~\cite{tan2024large}, effectively supporting hypothesis generation and validation processes~\cite{shen2022towards,wu2023visual}. Retrieval-augmented reasoning frameworks extend these capabilities further by dynamically integrating external knowledge during reasoning, thus enriching the comprehensiveness of exploratory analysis~\cite{piktus2021web}. Multi-agent systems are also proposed to collaboratively solve real-world data science challenges~\cite{li2024autokaggle}.

\vspace{.05in}
\noindent\textbf{Interactive AI Systems.}
Scaling reasoning steps and context length transforms interactive AI systems by significantly improving their adaptability and context-awareness. Expanded reasoning capabilities enable dialogue agents to maintain coherent and informative long-term interactions, effectively integrating historical context and external knowledge~\cite{bang2015example,friedman2023leveraging}. Multi-agent systems leverage iterative refinement and structured verification among specialized reasoning agents, further enhancing accuracy and reducing errors such as hallucinations~\cite{fang2024multi}. Interactive AI environments such as LLM-based Cursor~\cite{devi2024ai} leverage LLMs' contextual reasoning to facilitate precise user interactions, enabling targeted queries and refined outputs.

\vspace{.05in}
\subsection{Science}
The scaling of LLMs has significantly benefited scientific domains, with medicine, finance, and disaster management emerging as prominent application areas.

\vspace{.05in}
\noindent\textbf{Medical Domain.}
The medical domain has experienced remarkable advances through scaled LLMs. Research demonstrates that increasing model size leads to enhanced medical reasoning capabilities, with performance on medical questions improving proportionally~\cite{ brin2023large, lievin2024can, lu2025med}. This pattern extends to diagnostic reasoning~\cite{goh2024large, savage2024diagnostic}, where larger models can identify complex disease progression patterns that smaller models miss~\cite{zamora2024towards, garcia2024step}.
Multi-round reasoning approaches such as CoT have demonstrated exceptional effectiveness in medical diagnosis~\cite{weng2024large, liu2024medcot}, with additional reasoning steps yielding more accurate diagnoses~\cite{huang2025o1, chen2024huatuogpt} by enabling consideration of alternative explanations and confounding factors.
RAG techniques enhance medical question answering, with performance improving as the number of retrieved snippets increases~\cite{xiong2024benchmarking}.
Many-shot ICL shows particular efficacy for drug design tasks, with performance scaling with the number of examples provided~\cite{moayedpour2024many}.
Additionally, multi-agent LLM frameworks that simulate medical team consultations have demonstrated superior diagnostic accuracy, with specialized agents collaborating on complex cases to outperform single LLMs when benchmarked against gold-standard diagnoses~\cite{fan2024ai, kim2024mdagents}.

\vspace{.05in}
\noindent\textbf{Finance.}
Financial applications demonstrate improved performance with large-scale LLMs. Studies indicate that fine-tuned large-scale LLMs substantially outperform smaller alternatives~\cite{kalluri2024scalable}, with performance scaling with model size~\cite{qian2025fino1, li2024investorbench}.
The multi-step reasoning capabilities of scaled LLMs prove particularly valuable for complex financial analysis, significantly outperforming direct approaches~\cite{zhu2024tat, qian2025fino1}.
Financial sentiment analysis benefits from increased numbers of examples in many-shot ICL scenarios~\cite{agarwal2024many, wei2025large}.
RAG-based approaches incorporating banking webpages and policy guides improve question-answering performance, with results scaling with the number of retrieved documents~\cite{zhao2024optimizing}.
Multi-agent debate frameworks yield promising results in investment and trading decision scenarios~\cite{yu2024fincon, yu2024finmem, xiao2024tradingagents}, with specialized agents covering distinct functions outperforming single-agent approaches.

\vspace{.05in}
\noindent\textbf{Disaster Management.}
Disaster management has undergone substantial transformation through large-scale LLMs~\cite{lei2025harnessing}. Social media text classification for disaster types has improved significantly through LLM fine-tuning compared to traditional machine learning methods~\cite{yin2024crisissense, dos2024identifying}. The in-context learning capabilities of large-scale LLMs enable context-aware disaster applications including conversational agents for disaster-related queries and situational analysis~\cite{otal2024llm, rawat2024disasterqa}.
Large-scale disaster knowledge graphs enhance in-context learning through retrieval augmentation, enabling LLMs to generate more informative and less hallucinated responses~\cite{chen2024enhancing, xia2024question}.
For high-stakes disaster-related decision-making, multi-agent LLM approaches have been effectively deployed to facilitate adaptive and collaborative decision processes~\cite{dolant2025agentic, tran2025multi}, largely outperforming a single LLM.

\vspace{.05in}

\section{Future Directions}

\vspace{.05in}
\noindent\textbf{Efficiency in Scalable Reasoning.} Scaling reasoning capability in LLMs enhances their ability to solve complex problems but also increases response length, making it inefficient for simpler tasks. However, current LLMs apply uniform reasoning effort across all queries, leading to unnecessary computational overhead. A key direction for improvement is adaptive reasoning frameworks, where models dynamically adjust the depth of reasoning based on task difficulty~\cite{zhang2024generative, wei2025survey}. For example,  ``Proposer-Verifier'' framework~\cite{snell2024scaling} offers a promising approach by generating multiple candidate solutions and selecting the most reliable one through verification, reducing redundant reasoning steps while maintaining accuracy. However, achieving dynamic computation allocation requires robust uncertainty estimation, ensuring that models allocate resources efficiently without excessive overhead.

Another challenge is balancing search-based reasoning methods with computational cost. Approaches like ToT and Monte Carlo search refine reasoning iteratively but incur significant compute overhead. Selective pruning strategies that eliminate irrelevant reasoning paths while maintaining solution integrity could help optimize performance~\cite{xie2024monte}. Additionally, RL-based multi-step reasoning faces credit assignment issues, where sparse rewards make optimizing intermediate reasoning steps difficult~\cite{kumar2025llm}. Future work should explore hybrid reward models~\cite{shen2024improving} that combine process-based supervision (evaluating stepwise correctness) with some outcome-based rewards (final answer validation) to improve long-horizon reasoning stability and efficiency.

Beyond single-model scaling, collaborative multi-agent systems present a promising avenue for large-scale reasoning~\cite{le2024multi,owens2024multi}, but they also introduce significant coordination overhead. As the number of agents increases, computational redundancy and inefficient communication can slow down reasoning instead of improving it~\cite{guo2024large}. One approach to mitigate this is dynamic agent selection~\cite{liu2023dynamic}, where the system dynamically selects only the most relevant agents for a given reasoning task while discarding redundant ones. Another strategy is hierarchical multi-agent reasoning, where a smaller subset of expert agents handles complex queries, while simpler queries are resolved by lightweight, lower-cost agents. Additionally, inter-agent communication should be optimized through compressed latent representations rather than verbose token-based exchanges, further reducing computational overhead~\cite{zou2024genainet}. Future research should explore pruning and optimization techniques that enable multi-agent systems to scale efficiently without unnecessary computational waste, ensuring that reasoning is distributed optimally across agents.

\vspace{.05in}
\noindent \textbf{Inverse Scaling and Stability.}
Inverse scaling refers to the phenomenon where LLMs unexpectedly perform worse on certain tasks, contradicting standard scaling laws that predict consistent improvements with increased model size.  Lin et al.~\cite{lin2021truthfulqa} first observed this effect when evaluating LLMs such as GPT-2 and GPT-3 on truthfulness tasks, noting that common training objectives incentivize imitative falsehoods, where models produce false but high-likelihood responses due to patterns in their training distribution.  McKenzie et al.~\cite{mckenzie2024inverse} systematically analyzed different datasets exhibiting inverse scaling and identified key causes like solving distractor tasks instead of intended tasks.

While inverse scaling is widely observed, Wei et al.~\cite{wei2023inverse} challenge its universality, showing that some tasks previously exhibiting inverse scaling follow a U-shaped scaling trend—where performance initially declines with increasing model size but later recovers at even larger scales. This suggests that larger models can sometimes unlearn distractor tasks and correct their errors, emphasizing the importance of evaluating scaling trends beyond mid-sized models.

Since scaling laws were originally developed in the context of pretraining, they remain decoupled from downstream task performance, making it an open question of how to systematically predict and mitigate inverse scaling across different reasoning benchmarks. Additionally, challenges like reward hacking~\cite{amodei2016concrete}—where models exploit superficial signals rather than true reasoning improvements—necessitate adaptive reward models to maintain stability in multi-step reasoning.Future work should focus on developing predictive models for inverse scaling, refining adaptive fine-tuning methods, and leveraging world models for richer environmental feedback, ensuring that multi-step reasoning generalizes effectively across domains such as code generation, planning, question answering, and cross-lingual tasks.

\vspace{.05in}
\noindent \textbf{Security Risks in Scaled Reasoning Models.}
While CoT prompting enhances LLMs' ability to perform structured reasoning, it also introduces new security vulnerabilities, particularly backdoor attacks that manipulate the model’s reasoning process. BadChain~\cite{xiang2024badchain} exploits the model’s step-by-step reasoning by injecting backdoor reasoning steps, causing malicious alterations in the final response when a hidden trigger is present in the query. Similarly, H-CoT~\cite{kuo2025h} manipulates the model's internal reasoning pathways, hijacking its safety mechanisms to weaken its ability to detect harmful content. While defenses such as backdoor detection (CBD)~\cite{xiang2023cbd} and modified decoding strategies~\cite{jiang2025safechain} offer some protection, their effectiveness against novel attacks remains largely unexplored. This highlights the urgent need for more robust defenses capable of adapting to emerging threats.

Unlike CoT, RAG integrates external data sources, making them prone to data extraction attacks~\cite{cheng2024trojanrag}. Existing defenses primarily focus on retrieval corruption attacks~\cite{xiang2024certifiably,tan2024glue,zhou2025trustrag}, aiming to maintain performance, but data leakage prevention remains an underexplored area. For example, RAG-Thief demonstrates how attackers can extract scalable amounts of private data from proprietary retrieval databases \cite{jiang2024rag}. 
Beyond attacks on individual LLMs, the scaling of multi-agent reasoning systems introduces new attack surfaces. AgentPoison~\cite{chen2024agentpoison} specifically targets RAG-based and memory-augmented LLM agents, poisoning long-term memory or altering the knowledge base to induce faulty reasoning over time. As multi-agent LLM systems grow in scale, collusive behaviors among malicious agents present an even greater risk~\cite{yang2024watch}. BlockAgents proposes a blockchain-integrated framework for LLM-based cooperative multi-agent systems, mitigating Byzantine behaviors that arise from adversarial agents~\cite{chen2024blockagents}.

As AI adoption increases, the computational and environmental costs of inference also become a growing concern~\cite{luccioni2024power,samsi2023words, patterson2022carbon}. Large-scale LLMs demand significant energy resources on inference~\cite{patterson2022carbon}. This opens the door to a new form of attack, OverThink attack~\cite{kumar2025overthinking}, where an adversary intentionally inflates the number of reasoning tokens in an LLM’s response, drastically increasing financial and computational costs. As LLM reasoning continues to scale, deploying cost-effective safeguards against such attacks will become necessary for sustainable AI deployment.

%\vspace{-0.1in}
\section{Conclusion} % 1/4-page
% In this survey, we provided a comprehensive analysis of how scaling strategies influence reasoning capabilities in large language models. We examined four major dimensions: scaling in input sizes, reasoning steps, reasoning rounds, and model optimization, highlighting the methods, benefits, and challenges in each. While scaling improves LLM reasoning across many domains, it also introduces limitations such as computational inefficiency, instability, and new security risks. We emphasized emerging directions to address these issues, including adaptive computation, robust optimization, and safe multi-agent coordination. As LLMs continue to evolve, understanding and refining scalable reasoning will be key to building more capable, trustworthy, and efficient AI systems.

In this survey, we presented a comprehensive view of how different scaling strategies shape the reasoning capabilities of large language models. We organized the literature along four major dimensions: input information, reasoning steps, reasoning rounds, and model optimization, and discussed the key methods, benefits, trade-offs, and failure modes associated with each. Our analysis highlights that scaling can substantially improve LLM reasoning across a wide range of domains, but these gains are not uniform: they often come with increased computational cost, instability, diminishing returns, and emerging safety and security risks. We further outlined several promising directions for future research, including adaptive computation allocation, more robust and stable optimization, principled evaluation beyond final-answer accuracy, and safer multi-agent and human-LLM interaction. As LLMs continue to advance, a deeper understanding of how to scale reasoning effectively and responsibly will be essential for building AI systems that are not only more capable but also more efficient, reliable, and trustworthy.
%\end{document}  % This is where a 'short' article might terminate

%ACKNOWLEDGEMENTS are optional
\section{Acknowledgements}
This work is supported in part by the National Science Foundation (NSF) under grants IIS-2006844,
IIS-2144209, IIS-2223769, CNS-2154962, BCS-2228534, CMMI-2411248, ECCS-2143559, and
CPS-2313110; the Office of Naval Research (ONR) under grant N000142412636; and the Commonwealth Cyber Initiative (CCI) under grant VV1Q24-011.

\bibliographystyle{abbrv}
\bibliography{refs}  % sigproc.bib is the name of the Bibliography in this case
% ACM needs 'a single self-contained file'!
% SIGKDD: balancing columns messes up the footers: Sunita Sarawagi, Jan 2000.
% \balancecolumns

% That's all folks!
\end{document}